\documentclass{svproc}
\usepackage{graphicx}
\usepackage{amsmath}
\usepackage{mathtools}
\usepackage{multirow}

\usepackage{url}

\begin{document}
\mainmatter              
%
\title{Foreground object segmentation in RGB--D data implemented on GPU}
%
\titlerunning{FOS in RGB--D on GPU}  
%
\author{Piotr Janus\inst{1} \and Tomasz Kryjak\inst{1} \and Marek Gorgon}

\authorrunning{Piotr Janus et al.} 
%
\tocauthor{Piotr Janus, Tomasz Kryjak and Marek Gorgon}
%
\institute{AGH University of Science and Technology, \\ Faculty of Electrical Engineering, Automatics, Computer Science and Biomedical Engineering, Krakow, Poland. \\ e-mail: \{piojanus, tomasz.kryjak, mago\}@agh.edu.pl}

\maketitle              

\begin{abstract}

This paper presents a~GPU implementation of two foreground object segmentation algorithms: Gaussian Mixture Model (GMM) and Pixel Based Adaptive Segmenter (PBAS) modified for RGB--D data support.
The simultaneous use of colour (RGB) and depth (D) data  allows to improve segmentation accuracy, especially in case of colour camouflage, illumination changes and occurrence of shadows.
Three GPUs were used to accelerate calculations: embedded NVIDIA Jetson TX2 (Maxwell architecture), mobile NVIDIA GeForce GTX 1050m (Pascal architecture) and efficient NVIDIA RTX 2070 (Turing architecture).
Segmentation accuracy comparable to previously published works was obtained.
Moreover, the use of a~GPU platform allowed to get real-time image processing.
In addition, the system has been adapted to work with two RGB--D sensors: RealSense D415 and D435 from Intel.


\end{abstract}


\section{Introduction}
\label{sec:introduction}

Foreground object segmentation is one of the most important components of modern AVSS (\textit{Advanced Video Surveillance Systems}). 
It can be used in a~variety of vision systems such as object detection and tracking, as well as human behaviour analysis. 
Moreover, it is a~key element of applications like abandoned luggage detection or forbidden zone protection \cite{Guler_2016}.



The simplest group of foreground object detection algorithms is based on subtracting subsequent frames from a~video sequence. 
More advanced approaches involve the so-called background modelling. 
For each pixel, a dedicated model is assigned that describes the background appearance in a~given location. Then, depending on the used algorithm, the new pixel value is compared to the background model and classified as foreground, background and sometimes also as shadow. 
The model is usually updated to incorporate changes in the scene like slow or fast light variations and movement of background objects e.g. a~chair. 
The paper \cite{Bouwmans_2014} provides a~complete survey of the traditional and recent approaches in background modelling. 
Available resources, datasets and libraries are also presented.

However, some situations are difficult to handle by the proposed approach.
Examples involve: bootstrapping (model initialization), colour camouflage (object are similar to the background), illumination changes, intermittent motion (stopped or removed objects), background motion (like flowing water) and shadows -- a~more comprehensive discussion on this issue can be found in aforementioned paper and in \cite{Maddalena_2018}.

Some of the mentioned issues can be solved with the use of a~depth sensor.
Information about the scene geometry can be obtained in several ways.
The most straightforward is passive stereovision -- the use of two or more cameras and appropriate image algorithms allows to obtain a~3D representation of the scene.
Recently active sensors are gaining more and more attention: LiDAR, Time-of-Flight (ToF) cameras, structured light 3D scanners or  active IR (infrared) stereo. 
The last technology uses an IR emitter and one or two IR cameras.
The emitter displays an irregular pattern of dots.
Then the IR camera registers the infrared light reflected from the subjects.
Finally, the use od advanced image processing allows to estimate the depth map.
This approach is used in Microsoft Kinect (mono IR) and Intel RealSense (stereo IR) devices. 

On the other hands the use of depth information causes problems in specific situation like: depth camouflage (object close to the background), depth shadows, transparent or semi-transparent materials (like windows), out of sensor range -- a more detailed discussion can be found in \cite{Maddalena_2018}.
Therefore the majority of approaches involve combine colour (RGB) with depth (D) data. 
This type of image is usually called RGB--D (or RGBD).






In this paper two commonly used foreground segmentation algorithms Gaussian Mixture Model (GMM) and Pixel--Based--Adaptive--Segmenter (PBAS)  have been modified to include depth information. 
We used the \textit{Intel RealSense} D415 and D435 sensors for colour and depth image acquisition.
<<<<<<< HEAD
Three computing platforms were considered NVIDIA Jetson TX2 (embedded), NVIDIA GeForce GTX 1050m (mobile) and 
NVIDIA RTX 2070 (high-end).
GPU acceleration allowed to obtain real-time RGD--D data processing.
Moreover, we evaluated our approach on a~commonly used and publicly available dataset.

The reminder of this paper is organized as follows. 
In Section \ref{sec:prev_work} previous work related to use of RGB--D sensor for foreground object segmentation are briefly discussed and papers concerning GMM and PBAS acceleration using GPU are also presented. 
Section \ref{sec:algorithms} describes the proposed versions of GMM and PBAS methods. 
In Section \ref{sec:hw_implementation} the designed heterogeneous system is presented. 
The evaluation of the proposed algorithms is discussed in Section \ref{sec:evaluation}. 
The paper ends with a conclusion and discussion of future research directions.


\section{Previous work}
\label{sec:prev_work}


Over the years, several solutions for foreground object segmentation with the use of a~RGB--D sensor have been proposed. 
An excellent and quite recent (2018) review is presented in \cite{Maddalena_2018}. 
Here we limit our discussion to papers that use algorithms comparable with our approach i.e. GMM and PBAS.

One on the first works on using RGB--D data in foreground segmentation was \cite{Gordon_1999}.
The approach was originally applied to stereovision data.
It was based on the Mixture of Gaussian (MoG, also known as Gaussian Mixture Models) concept.
The authors assumed that colour and depth features are independent.
They also divided the depth data into ``valid'' and ``invalid''.
In the first case, the depth was used to estimate the background, as usually it is behind the foreground (an exception are occlusions).
In the second case, the typical colour-based MoG algorithm was used.
During segmentation the depth data was used to influence the colour-based matching criterion.
For reliable depth data, the criterion was relaxed to avoid camouflage errors.
In the other case, the criterion was harder, to avoid segmentation errors due to shadows and illumination changes.
No information about processing time was provided. 

The MoG approach was also described in \cite{Stormer_2010}.
It was originally applied to data obtained by a~ToF sensor, which provided depth data and a~near infrared image.
In contrast to \cite{Gordon_1999}, the authors used two separate models for depth and IR image and obtained two foreground masks.
The final segmentation was based on the fusion of these masks with additional information about depth gradient to separate overlapping foreground objects.  
No information about processing time was provided. 

Another GMM based algorithm adapted to work with a~RGB--D sensor was proposed in \cite{Song_2014}.
The authors used two separate models and combined their output to obtain the final segmentation result.
The evaluation was done on sequences recorded by the authors.
It showed that the proposed approach works well in case of colour camouflage.
No information about processing time was provided. 

A~different algorithm -- ViBE (Visual Background Extractor) -- was used in the work \cite{Leens_2009}.
It was applied to ToF data and two separate models were used.
Moreover motion information was also included.
The obtained foreground masks were combined and post-processed with morphological operations.
No information about processing time was provided. 

In the paper \cite{Minematsu_2017} an algorithm, named SCAD, based on ViBE and combination of colour, texture and depth information was proposed. 
The final segmentation was obtained using graph cuts.
The solution was implemented in C++ as a single thread application on a~Intel Xeon @ 3.7 GHz with 32 GB RAM.
The system processed, on average, $640 \times 480 @ 1.95$ frame per second.

A~similar approach was presented in \cite{Zhou_2017}.
The authors fused segmentations results from two ViBE models: for colour and depth.
The solution was implemented in C++/OpenCV. 
On a~Intel Core Duo 2 CPU E7500 platform with 2.00 GB of RAM, a $640 \times 480 @ 30$ fps performance was obtained.

There are several published papers on GPU acceleration of foreground object detection algorithms like: GMM, ViBE and PBAS.
In \cite{Pham_2010} GMM was implemented on a~low end GPU -- GeForce 9600GT.
This allowed to process up to 50 HD frames per second.
In \cite{Karahan_2015} a NVIDIA Tesla K20 GPU was used to accelerate PBAS.
For a~$320 \times 240$ video 646 fps was reported.
In the paper \cite{Qin_2015}, a~GPU implementation of the ViBE algorithm was presented. 
The algorithm was tested on a~PC equipped with \textit{Intel Core Quad Q8400} and \textit{Nvidia GTX 650Ti}.
For a~$960x540$ video stream resolution the achieved performance is $1.8$ fps for CPU  and $26$ fps for the GPU implementation.  
In \cite{Kumar_2013} another variant of the GMM algorithm with connected component labelling and morphological operations for post--processing is described. 
The authors presented a~PU implementation which achieves significant speed-ups of 15 times for the GMM algorithm comparing to \textit{Intel Xeon processor}. 
The proposed system is able to process 22.3 frames per second for HD video stream.

\section{The considered algorithms}
\label{sec:algorithms}


In this research, the authors implemented two different background subtraction algorithms. 
The standard RGB version were modified to take benefit for depth data.
The fist algorithm is an extended version of \textit{Gaussian Mixture Models} \cite{Stauffer_1999}, while the second one is a~modification of \textit{Pixel Based Adaptive Segmenter} \cite{Hofmann_2012}. 
Both are similar regarding the background model concept. 
It is independent for each pixel and dynamically updated after every frame. 
In the following subsections a~detailed description of both methods is presented. 


\subsection{GMM algorithm with RGB--D data}
\label{subsec:gmm_rgbd}

Gaussian Mixture Models (GMM, also known as Mixture of Gaussians (MOG)) \cite{Stauffer_1999} is one of the most commonly used method for background modelling. 
In this approach each pixel is modelled by $k$ Gaussian distributions characterized by three parameters ($\omega$, $\mu$, $\sigma^2$), where $\omega$ is the normalized weight (range 0--1) of the Gaussian distribution, $\mu$ is the means vector of each colour component of a~particular pixel -- ($r_{mean}$, $g_{mean}$, $b_{mean}$) is case of RGB, and  $\sigma^2$ is the variance of given Gaussian distribution -- a~single value is used for each colour component.
Usually it is assumed that RGB components are independent, which allows to use three $\sigma^2$ values instead of a~covariance matrix. 
In this work, a~version partially based on \cite{Song_2014} and the open source image processing library OpenCV was applied. 
As a~detailed description of the method is quite long and available in multiple research papers (starting with \cite{Stauffer_1999}), we only discuss the adaptation to RGB--D data.

There are two options when applying a~RGB background model to RGB--D data.
The first is to incorporate the depth data into the model.
In the context of GMM this results in extending the $\mu$, $\sigma^2$ parameters.
The second is to use two different models and then combine the segmentation results.
In the presented research we followed the second approach. 
During our experiments we found out that such an approach provides better protection against noise from depth image and illumination changes. 

\begin{figure}[!t]
	\begin{center}
		\includegraphics[scale=0.45]{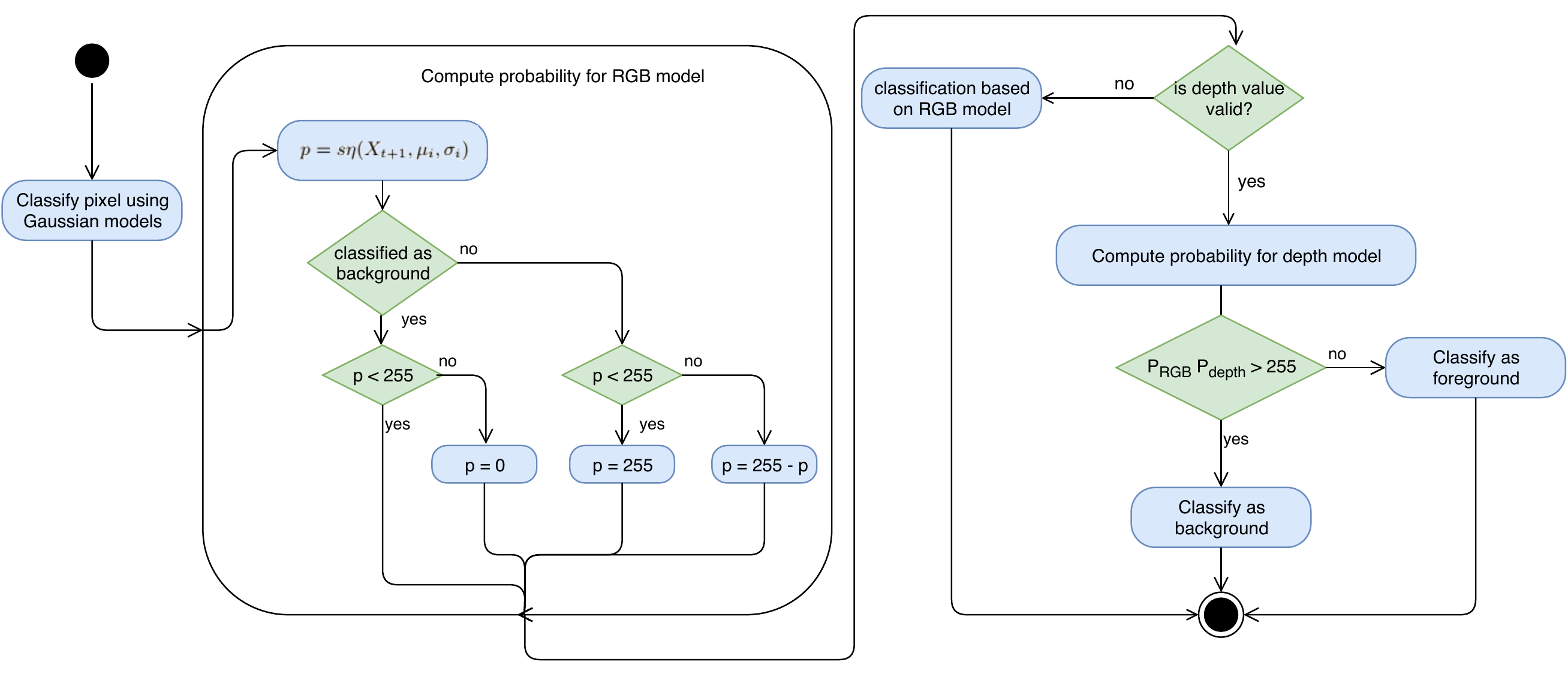}
		\caption{GMM -- computing probability and classification}
		\label{fig:gmm_alg}
	\end{center}
\end{figure}

The classification procedure is based on a~probability density function, which depends on the pixel value $X_t$ at time $t$:
\begin{equation}
\eta (X_t, \mu, \sigma) = \frac{1}{2\pi\sigma} e^{-\frac{d(X_t, \mu)^2}{2\sigma}} 
\label{equ:gmm_density}
\end{equation}
This operation is presented in Figure \ref{fig:gmm_alg}.
The $s$ parameter is used for scaling the probability density and its default value is $10000$. 
The process of computing the probability factor is the same for both models. 
Then the product of two values is computed and final classification is made according to the diagram. In addition depth model is considered only if depth value obtained from sensor is valid (greater than 0) otherwise only RGB classification is performed. In final implementation the following algorithm parameters are used: number of gaussians -- $7$ for RGB model and $3$ for depth, learning rate -- 0.001, gaussian parameters are represented as float number on 64 bit machine.

\subsection{PBAS algorithm with RGB--D data}
\label{subsec:pbas_rgbd}

The Pixel Based Adaptive Segmenter (PBAS) algorithm \cite{Hofmann_2012} is an extension of the Visual Background Extractor method proposed in \cite{Barnich_2009}.
Both algorithm use a~similar background model, however PBAS involves a~more advanced foreground classification and model update procedure.
As a~detailed description of the method is quite long and available in \cite{Hofmann_2012}, we only discuss the adaptation to RGB--D data.

In the PBAS method the background model is composed of two parts.
The first one is a buffer of $N$ samples from the analysed video sequence. 
In our approach a~single sample consists both of a RGB and a~depth value.
Since in segmentation and model update each component is processed separately, the addition of depth data is straightforward. Moreover, as in case of GMM algorithm, depth data is utilized only if depth value from sensor is valid, otherwise classification is based only on RGB model. 
In final implementation a~model containing $20$  samples is used.

\section{Algorithms implementation}
\label{sec:hw_implementation}


\subsection{Hardware setup}
\label{subsec:hw_used}

We used two RGB--D video sources.
To compare our solution with other approaches we used a~publicly available dataset -- \cite{SBM_RGBD_2017}.  Also to demonstrate a~vision system working in real-time we used Intel RealSense Depth Cameras D415 and D435.
Both sensors provide a~Full HD resolution ($1920 \times 1080$) for the RGB image and HD resolution ($1280 \times 720$) for the depth map. 
They are able to distinguish objects in range 10 centimetres to 10 meters from the camera~lens. 
A~well known alternative to \textit{RealSense} sensors is Microsoft Kinect. 
It has been used for a~long time as an entry level device for RGB--D image analysis, also in the used dataset \cite{SBM_RGBD_2017}. 
Unfortunately it was discontinued by Microsoft in 2017 and it is no longer available on the market. 
Moreover, its technical specification is significantly inferior than \textit{Intel RealSense} sensors, as the maximum resolution for RGB camera and depth map is only $640 \times 480 @ 30Hz$. 
Nowadays D415 and D435 devices are affordable, provide a~reasonable price to value ratio and are used in a~rage of robotic applications like drone navigation \cite{Campos_2019}. 



For GPU implementation three different platforms were used. 
The first one was NVIDIA Jetson TX2 -- an embedded GPU, equipped with a~64-bits ARM Cortex A57 CPU and a~NVIDIA Maxwell GPU with 256 CUDA cores.
The second platform was a~laptop with a~Intel Core i7--7700HQ (4 cores/8 threads @ 2.8 GHz) CPU and a~NVIDIA GeForce GTX 1050m GPU based on Pascal architecture. 
The third was a~PC equipped with a~Intel Core i7--9700k (8 cores @ 4.5 GHz) CPU and a~NVIDIA RTX 2070 GPU (Touring architecture).


To accelerate RGB--D data processing the CUDA (\textit{Compute Unified Device Architecture}) platform was used. 
It is developed by NVIDIA and can be used only with GPUs based on their architecture. 
The CUDA itself is a~parallel computing platform with an API, which allows to use NVIDIA graphics processing unit for general purpose computing. 
Thanks to this approach, it is straightforward to port our implementation to different computing platforms like laptops, PCs or embedded GPUs. 
The only requirement is the use of an NVIDIA GPU. 



\subsection{Application architecture}
\label{subsec:architecture}

\begin{figure}[!t]
	\begin{center}
		\includegraphics[scale=0.70]{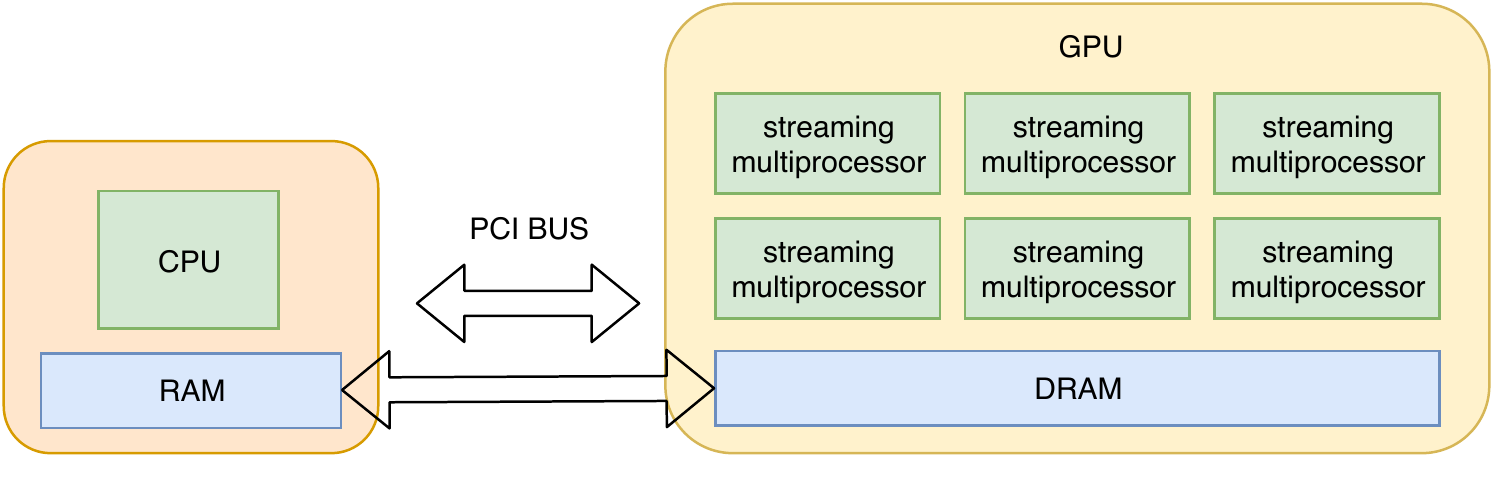}
		\caption{Communication between host (CPU) and GPU}
		\label{fig:cpu_host}
	\end{center}
\end{figure}

The distribution of computing tasks between CPU and GPU is one of the key element when implementing a~heterogeneous system. 
The host (CPU) is responsible for image acquisition (from the sensor or hard drive) and copying image data to the shared GPU memory (DRAM).
Moreover, the memory allocation for the background model is also done by the host. 
The communication between host and GPU is done over a~PCI bus. 
A~general overview of this architecture is shown in Figure \ref{fig:cpu_host}.    


\begin{figure}[!t]
	\begin{center}
		\includegraphics[scale=0.70]{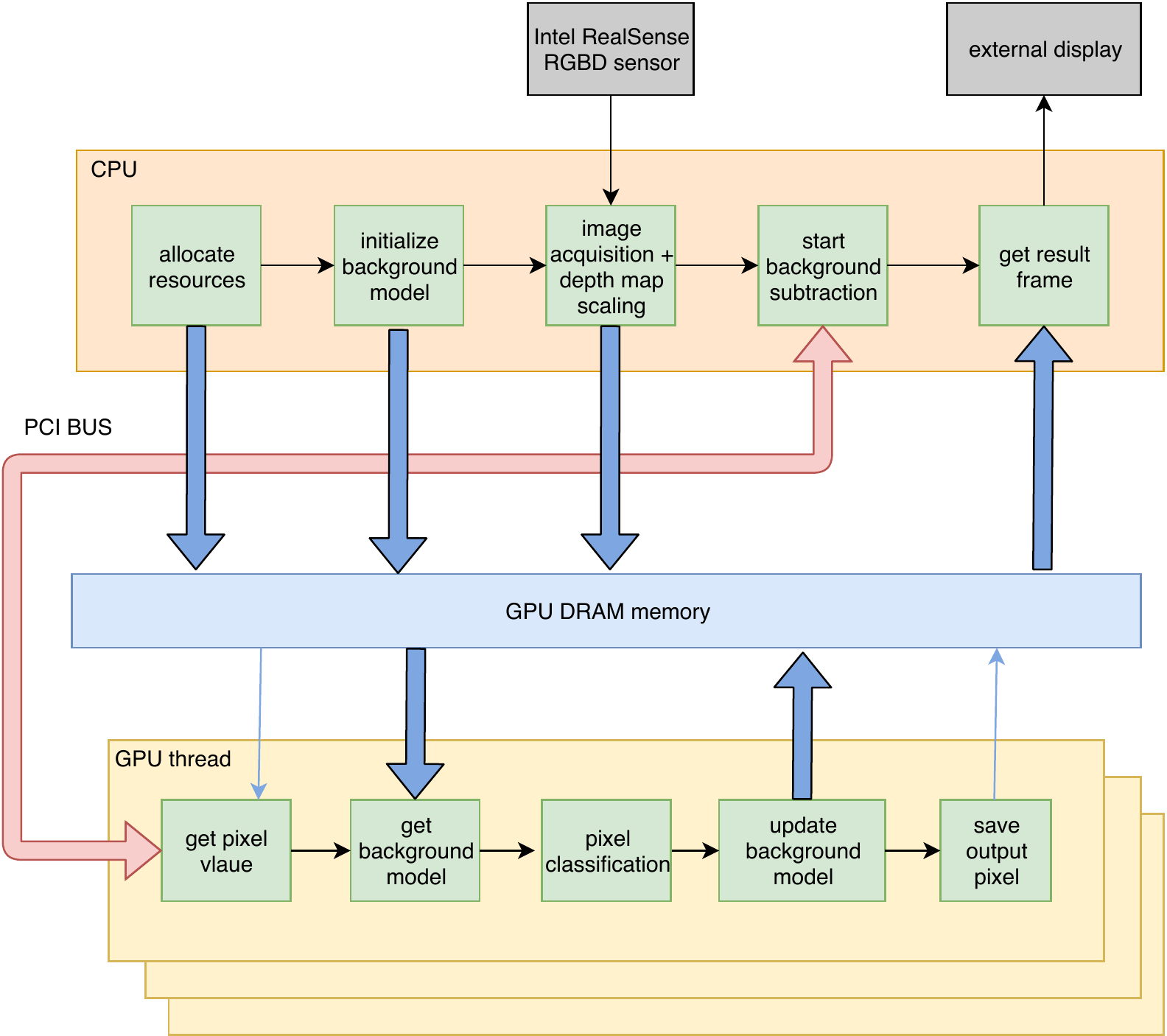}
		\caption{Exchanging data between host and GPU -- flow diagram}
		\label{fig:gpu_flow}
	\end{center}
\end{figure}

The GPU architecture allows to process each pixel  parallel, independently, using separate threads.
The output of the system is a~binary mask containing foreground objects. 
This mask has to be copied from the GPU DRAM memory to CPU RAM and then forwarded to an external display or stored on a~hard drive. 
The flow of exchanging data and executing operations on the CPU and GPU sides is presented in Figure \ref{fig:gpu_flow}. 




The CPU part was implemented in \textit{C++} language. 
Image acquisition was done with the use of dedicated SDK provided by \textit{Intel} for \textit{RealSense} RGB--D sensors. 
It allows to acquire both RGB images and depth maps in real-time with a~maximum frequency limited to 30 fps. 
The depth map is received as a 16-bit unsigned integer (a~bigger number means a~greater distance from the camera), so it needs to be converted to 8-bit per pixel format for compatibility with the RGB background model based on 8-bit numbers.
As it was mentioned before, according to the hardware specification, the minimal range of the depth sensor is 10 cm, while the maximum is 10 meters. 
After rescaling from the range 0--65535 to 0--255, the depth accuracy will be about 4~cm, which is enough for the considered vision system. 
Finally each pixel is represented by 32 bits, 8 bits for each colour component and depth.


\subsection{Performance}
\label{subsec:performance}

\begin{figure}[!t]
	\begin{center}
		\includegraphics[scale=0.12]{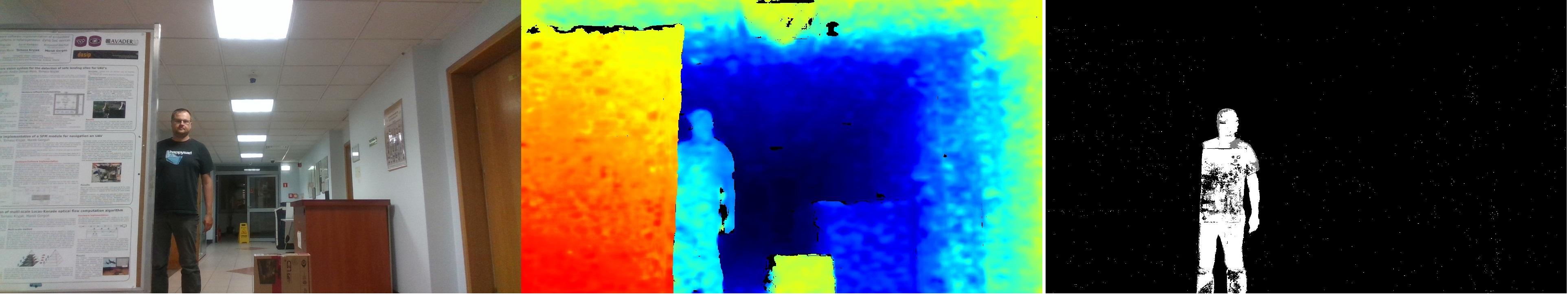}
		\caption{Exemplary RGB, depth input and segmentation output for the GMM algorithm}
		\label{fig:gmm_example}
	\end{center}
\end{figure}

Computing performance on particular GPUs has been measured for different resolutions: 480p/480p, 720p/480p, 720p/720p, 1080p/720p, where values represent RGB camera and depth resolution respectively.
Results are presented in Table \ref{tab:performance_res} and a~sample RGB image, depth map and segmentation output shown in Figure~\ref{fig:gmm_example}.
It should be noted, that similar results were obtained for the GMM and PBAS methods.


\begin{table}[t!]
	\caption{Performance}
	\centering
	\begin{tabular}{|c|c|c|c|c|c|}
		\hline
		&     480p/480p  & 720p/480p & 720p/720p & 1080p/720p \\ \hline
		Jetson TX2  & 28fps & 11fps & 9fps & 6fps  \\ \hline
		i7 7700hq + GTX 1050m   & 30fps & 18fps & 16fps & 10fps     \\ \hline
		i7 9700k + RTX 2070 & 30fps & 30fps &  30fps & 30fps    \\ \hline
	\end{tabular}
	\label{tab:performance_res}
\end{table}


Analysing the obtained results, it can be noticed that only the RTX 2070 GPU is able to provide real-time processing for all considered resolutions.
It should be emphasized that for the RealSense sensor, both the RGB image and the depth map are acquired 30 times per second.
Thus, 30 fps is for the considered system the maximum value.

In the case of  mobile GTX 1050m and embedded Jetson TX2 GPUs, only for 480p resolution it is possible to obtain real-time image processing.
However, it is worth noting that these are definitely more energy efficient platforms than RTX GPUSs with Touring architecture.
The maximum power consumption for the considered platforms is 7.5W, 75W and 215W respectively.

A decrease in performance can be seen when different resolutions for the depth map and image are used.
This is due to the depth map upscaling to the same resolution as the RGB image.




\section{Evaluation}
\label{sec:evaluation}

Two experiments were performed to test the implemented algorithms.
In the first one, short videos were recorded using the Intel RealSense D435 sensor.
They contained situations when the object's colour was similar to the background (colour camouflage) and when the object was close to the background (depth camouflage).
For each recording a~ground truth has been prepared.
In the second experiment, test sequences from the SBM RGBD \cite{SBM_RGBD_2017} dataset were used. 
This allowed to compare the proposed methods with those described in the literature \cite{Maddalena_2018}.


\begin{table}[t]
	\caption{Evaluation results on sequences registered with RealSense}
	\label{tab:evaluation_res_RS}
	\centering
	\begin{tabular}{|c|c|c|c|c|}
		\hline 
		Algorithm & PWC & FNR & FPR & Si \\
		\hline
		GMM & 5.85 & 0.1287 & 0.0423 & 0.51   \\
		\hline  
		GMM + RGBD & 4.21 & 0.0812 & 0.0323 & 0.61   \\
		\hline
		PBAS & 1.23 & 0.0398 & 0.0123 & 0.87  \\
		\hline
		PBAS + RGBD & 1.20 & 0.0289 & 0.113 & 0.89   \\
		\hline
	\end{tabular} 
\end{table}

A typical evaluation methodology was used.
Based on the comparison of the segmentation result and the ground-truth mask, the following factors were determined:
\begin {enumerate}
\item [TP]: the number of pixels correctly classified as the foreground (true positive),
\item [TN]: number of pixels correctly classified as background (true negative),
\item [FN]: number of pixels incorrectly classified as a background (false negative),
\item [FP]: the number of pixels incorrectly classified as the foreground (false positive).
\end{enumerate}


Then four quality indicators were determined:
 
\begin{enumerate}
	\item Percentage of Wrong Classifications (PWC) : $100(FN + FP)/(TP + FN + FP + TN)$
	\item False Negative Rate (FNR): $FN/(TP + FN)$
	\item False Positive Rate (FPR): $FP/(FP + TN)$
	\item Similarity (Si) : $TP/(TP + FP+ FN)$
\end{enumerate}

The test results for the sequences registered with the RealSense sensor are presented in Table \ref{tab:evaluation_res_RS}.
On their basis, it can be concluded that in the case of the GMM algorithm, the obtained results are definitely better when using the RGB--D sensor.
For the PBAS method, adding depth data only slightly improved the segmentation.


Four sequences from the SBM RGBD dataset belonging to different categories were used: illumination changes, colour camouflage, depth camouflage and shadows.
The obtained results are presented in Table \ref{tab:evaluation_res} and compared with previously proposed solutions \textit{MoG4D} \cite{Gordon_1999}, \textit{ViBeRGB+D} \cite{Leens_2009}, \textit{MoGRGB+D} \cite{Stormer_2010}.
The selected algorithms are comparable to ours i.e. with similar background model and computational complexity.
It should also be noted that the SBM RGBD set contains more sequences. 
However, only for the above mentioned the evaluation results are presented in \cite{Maddalena_2018} for the  \textit{MoG4D}, \textit{ViBeRGB+D} and \textit{MoGRGB+D} algorithms.


\begin{table}[t]
\caption{Evaluation on the SMB RGBD dataset}
\label{tab:evaluation_res}
\centering
\begin{tabular}{|c|c|c|c|c|c|}
  \hline 
  Sequence & Method & PWC & FNR & FPR & Si \\
  \hline
  \multirow{7}{*}{Illumination changes} & GMM & 4.49 & 0.0248 & 0.4783 & 0.73   \\
  \cline{2-6}  
  & PBAS & 3.75 & 0.0212 & 0.0412 & 0.75   \\
  \cline{2-6}
  & GMM + RGBD & 3.60 & 0.0183 & 0.0376 & 0.78  \\
  \cline{2-6}
  & PBAS + RGBD & 3.32 & 0.0131 & 0.0319 & 0.80   \\
  \cline{2-6}
  & MoG4D & 1.93 & 0.0063 & 0.0209 & 0.79  \\
  \cline{2-6}
  & ViBeRGB+D & 12.39 & 0.0065 & 0.1385 & 0.44  \\
  \cline{2-6}  
  & MoGRGB+D & 2.03 & 0.1701 & 0.0016 & 0.79  \\
  \hline
  \multirow{7}{*}{Color Camouflage} & GMM & 18.89 & 0.8001 & 0.0154 & 0.19   \\
  \cline{2-6}  
  & PBAS & 17.87 & 0.7702 & 0.0112 & 0.19   \\
  \cline{2-6}
  & GMM + RGBD & 10.01 & 0.0160 & 0.0124 & 0.72  \\
  \cline{2-6}
  & PBAS + RGBD & 9.02 & 0.0125 & 0.0930 & 0.76   \\
  \cline{2-6}
  & MoG4D & 3.49 & 0.0038 & 0.0613 & 0.91  \\
  \cline{2-6}
  & ViBeRGB+D & 6.94 & 0.0017 & 0.1269 & 0.81  \\
  \cline{2-6}  
  & MoGRGB+D & 38.47 & 0.8287 & 0.0075 & 0.22  \\
  \hline
  \multirow{7}{*}{Depth Camouflage} & GMM & 7.24 & 0.5108 & 0.4734 & 0.26   \\
  \cline{2-6}  
  & PBAS & 6.91 & 0.4912 & 0.0438 & 0.31   \\
  \cline{2-6}
  & GMM + RGBD & 7.22 & 0.5001 & 0.0465 & 0.27  \\
  \cline{2-6}
  & PBAS + RGBD & 6.89 & 0.4832 & 0.0435 & 0.32   \\
  \cline{2-6}
  & MoG4D & 2.11 & 0.1525 & 0.0131 & 0.61  \\
  \cline{2-6}
  & ViBeRGB+D & 9.31 & 0.0548 & 0.0955 & 0.30  \\
  \cline{2-6}  
  & MoGRGB+D & 3.57 & 0.6087 & 0.0009 & 0.32  \\
  \hline      
  \multirow{7}{*}{Shadows} & GMM & 14.60 & 0.6754 & 0.0603 & 0.23   \\
  \cline{2-6}  
  & PBAS & 10.24 & 0.33 & 0.04 & 0.32   \\
  \cline{2-6}
  & GMM + RGBD & 9.12 & 0.1409 & 0.0412 & 0.43  \\
  \cline{2-6}
  & PBAS + RGBD & 8.99 & 0.1023 & 0.0298 & 0.46   \\
  \cline{2-6}
  & MoG4D & 3.94 & 0.0059 & 0.0450 & 0.77  \\
  \cline{2-6}
  & ViBeRGB+D & 7.15 & 0.0001 & 0.0834 & 0.66  \\
  \cline{2-6}  
  & MoGRGB+D & 3.43 & 0.2351 & 0.0008 & 0.75  \\
  \hline
\end{tabular} 
\end{table}

Analysing the results, we can conclude that using depth map information allows to obtained are better results in each category.
The biggest difference can be seen in the case of colour camouflage.
As expected, in this situation depth information gives the greatest benefits.
A clear improvement was also seen in the sequence containing a~lot of shadows.
In the case of changes in lighting and depth camouflage, the benefits of using depth maps were not so impressive.

We noticed, that the PBAS algorithm allows to obtain, depending on the test sequence, slightly or clearly better results than the GMM method.
Among the analysed methods, the best results are obtained by the \textit{MoG4D} algorithm, most likely due to a~more advanced method of analysing incorrect depth values.
The methods implemented in this paper obtain comparable results to \textit{MoGRGB+D} and \textit{ViBERGB+D}.

\section{Conclusion}

In this paper the implementation of GMM and PBAS algorithms adapted to RGB--D data has been presented.
In both cases, hardware acceleration with CUDA parallel computing platform was used.
The system was launched on three GPUs with different levels of performance from embedded Jetson TX2, through GTX 1050m, ending with RTX 2070 with Touring architecture.
The last of the mentioned platforms allowed to obtain real-time processing for 1080p data (30 fps).
The performed evaluation showed that using the RGB--D sensor provides an increase in segmentation accuracy.
As expected, the largest improvement was reached for the ``colour camouflage'' case, when objects have similar colour to the background.

As part of future work, the proposed algorithms could be improved by adding more advanced fusion of RGB and depth data, as well as detection of static objects.
For example, the approach proposed in \cite{Kryjak_pbas_2014} could be applied.
In addition, it is worth to consider preparing a set of sequences registered with various sensors: Kinect (like SBM RGBD), RealSense, a stereo camera and a ToF sensor.
This would allow to compare different algorithms on different RGB--D data and evaluate which solution is best for foreground object segmentation.
Another research direction could be the acceleration of RGB--D algorithms using FPGA devices, as this could potentially allow real-time processing of a stream with a resolution of 1080p with significantly lower power consumption than the RTX 2070 GPU.

\end{document}